\documentclass{article}

\usepackage[final]{nips_2017}

\usepackage[utf8]{inputenc} 
\usepackage[T1]{fontenc}    
\usepackage{hyperref}       
\usepackage{url}            
\usepackage{booktabs}       
\usepackage{amsfonts}       
\usepackage{nicefrac}       
\usepackage{microtype}      
\usepackage{graphicx}
\usepackage{wrapfig}

\title{Synthetic Medical Images from Dual Generative Adversarial Networks
}

\author{
  John T. Guibas\textsuperscript{*} \\
   Student\\
  Henry M. Gunn High School\\
  Palo Alto, CA 94306 \\
  \texttt{jtgg01@gmail.com} \\
  \And
  Tejpal S. Virdi\textsuperscript{*} \\
   Student\\
  Henry M. Gunn High School\\
  Palo Alto, CA 94306 \\
  \texttt{tejpalv8@gmail.com} \\
  \And
  Peter S. Li \\
   Student\\
  Henry M. Gunn High School\\
  Palo Alto, CA 94306 \\
  \texttt{peter.s.li93@gmail.com} \\
}
\begin{document}

\maketitle

\begin{abstract}
  Currently there is strong interest in data-driven approaches to medical image classification. However, medical imaging data is scarce, expensive, and fraught with legal concerns regarding patient privacy. Typical consent forms only allow for patient data to be used in medical journals or education, meaning the majority of medical data is inaccessible for general public research. 
  We propose a novel, two-stage pipeline for generating synthetic medical images from a pair of generative adversarial networks, tested in practice on retinal fundi images. We develop a hierarchical generation process to divide the complex image generation task into two parts: geometry and photorealism.
  We hope researchers will use our pipeline to bring private medical data into the public domain, sparking growth in imaging tasks that have previously relied on the hand-tuning of models. We have begun this initiative through the development of SynthMed\footnote{https://synthmed.github.io}, an online repository for synthetic medical images.
  
\let\thefootnote\relax\footnotetext{*Equal contribution}

\end{abstract}

\section{\label{sec:level1}Introduction}
Computer-aided medical diagnosis is widely used by medical professionals to assist in the interpretation of medical images [1]. Recently, deep learning algorithms have shown the potential to perform at higher accuracy than professionals in certain medical image understanding tasks, such as segmentation and classification [2]. Along with accuracy, deep learning improves the efficiency of data analysis tremendously, due to its automated and computational nature. Since most medical data is produced in large volumes, and is often 3-dimensional (MRIs, CTs, etc.), it can be cumbersome and inefficient to annotate manually. 

There is strong interest in computer-aided medical diagnosis systems that rely on deep learning techniques [3]. However, due to proprietary and privacy reasons limiting data access [4], the development and advancement of these systems cannot be accelerated by public contributions. It is difficult for medical professionals to make most medical images public without patient consent [5]. In addition, the publicly available datasets often lack size and expert annotations, rendering them useless for the training of data-hungry neural networks. The design of these systems is therefore done exclusively by researchers that have access to private data, limiting the growth and potential of this field of research. 

In the last 10 years, many breakthroughs in deep learning attribute success to extensive public datasets such as ImageNet. The annual ImageNet competition decreased image recognition error rates from 28.2\% to 6.7\% [6] in the span of 4 years from 2010 to 2014. ImageNet required the work of almost 50,000 people to evaluate, sort, and annotate one billion candidate images [6]. This showcases that access to large and accurate datasets is extremely important for building accurate models. However, current research in the field of medical imaging has relied on hand-tuning models rather than addressing the underlying problem with data. We believe that a public dataset for medicine can spark exponential growth in imaging tasks.

We propose a novel pipeline for generating synthetic medical images, allowing for the production of a public and extensive dataset, free from privacy concerns. We put this into practice through SynthMed, a public repository for these images.

\section{\label{sec:level1}Related Works}
Researchers across a variety of disciplines have taken private data to the public domain using synthetic data. For example, the U.S. Census collects personally identifiable information (PII) such as occupation, education, income, and geographical data for the US population. Due to the natural specificity of the data, even if sources are de-identified and obfuscated, there is considerable risk of deanonymization [7]. This valuable data, which holds many potentially useful statistical correlations, is publicly unavailable because of privacy issues. Reiter, a researcher at Duke University, solved this privacy problem by generating synthetic business census data [8]. In 2011, they released the Synthetic Longitudinal Business Database [9], the first publicly available record-level database on business establishments. 

As seen in Reiter’s research, previous uses of synthetic data to bring private data to the public domain have been done solely with scalar quantities. With the growing power of data-driven computer vision techniques, this paper explores the idea of synthetic data for images. Recent developments of neural networks, specifically the generative adversarial network (GAN) [12], promise the possibility for more realistic image generation. However, images produced by a GAN may often still contain artifacts and noise, due to instabilities in locating a saddle point in the energy landscape. We address this issue by creating a novel image generation pipeline using a pair of GANs to promote increased stability. 

\section{Data}
We trained the GAN in Stage-I with retinal vessel segmentations from the DRIVE database [10]. DRIVE contains forty pairs of retinal fundi images and vessel segmentation masks manually labeled by two experts. The GAN in Stage-II was trained with segmentation masks, derived from a segmentation network, and corresponding photorealistic images from MESSIDOR [11]. We also used DRIVE to train a single GAN model to compare our results. One u-net was trained on DRIVE, and the other u-net was trained on 50 pairs of GAN-produced images.

\begin{figure}[!htb]
  \centering
  \includegraphics[width=200pt]{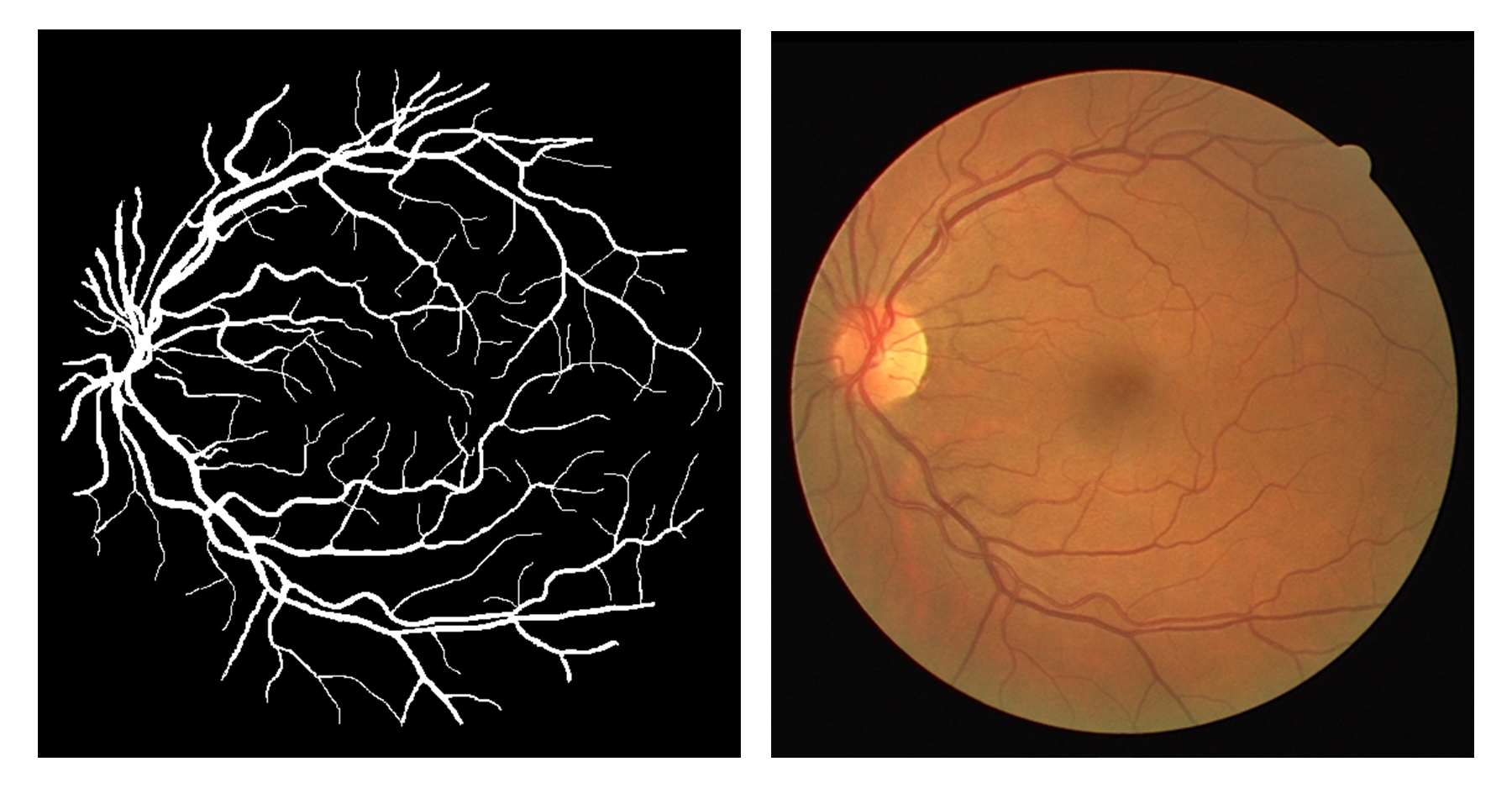}
  \caption{Example vessel tree segmentation mask and retina fundus image from DRIVE.}
  \label{fig:boat1}
\end{figure}

\section{General Pipeline}
To generate a high quality synthetic dataset, we propose the use of two GANs, breaking down the generation problem into two parts: 

\begin{enumerate}
    \item Stage-I GAN: Produce segmentation masks that represent the variable geometries of the dataset.
    \item Stage-II GAN: Translate the masks produced in Stage-I to photorealistic images.
\end{enumerate}

We illustrate the process with retinal fundi images.

\begin{figure}[!ht]
\includegraphics[width=\linewidth]{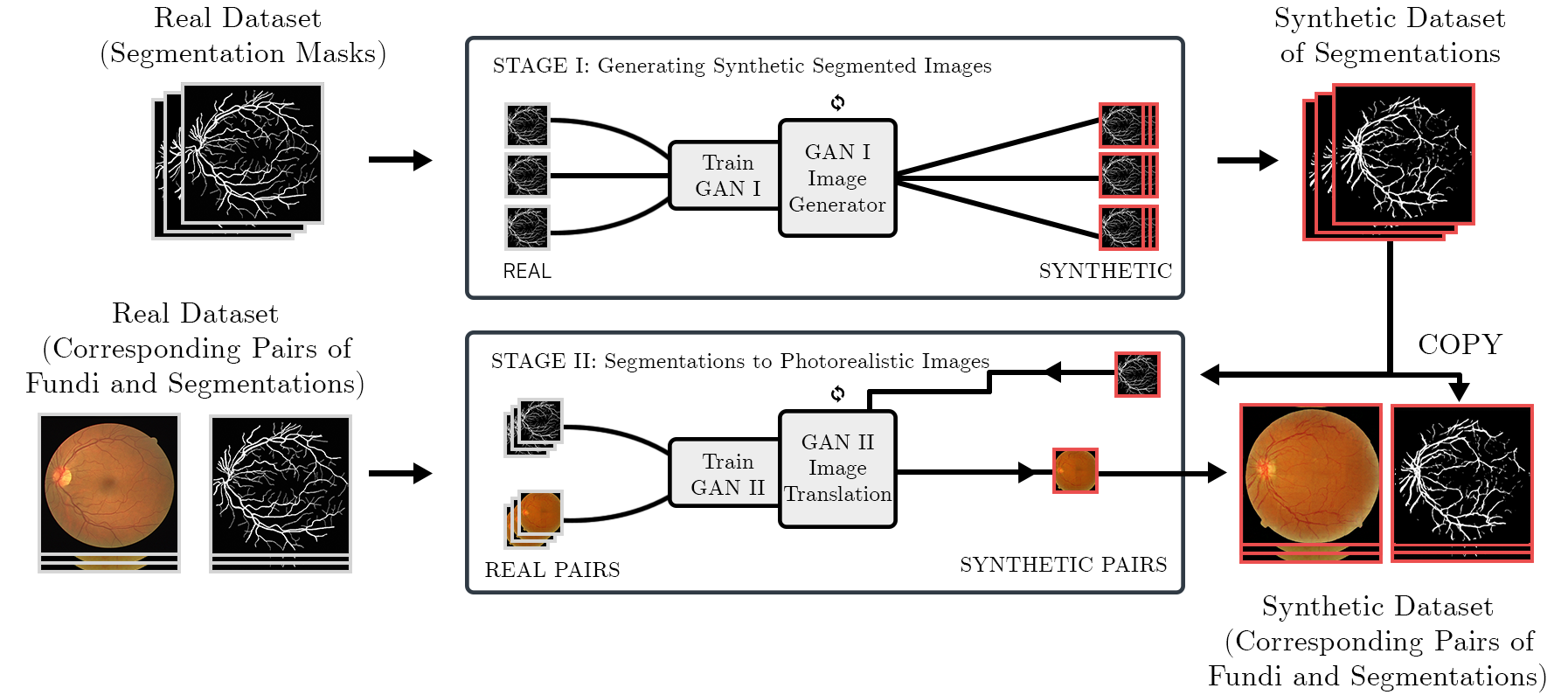}
\caption{Flowchart of proposed pipeline.}
\end{figure}

\section{Generative Adversarial Network}
The Generative Adversarial Network (GAN), as proposed by Goodfellow et al. [12] in June 2014, involves the competition between two models: the discriminator D and the generator G. D is a binary classifier that classifies the data produced by G as either part of the training set (realistic) or not (unrealistic). G minimizes its loss function by producing data that D will classify as real, as modeled by:

\smallskip

\[minmax(D,G)=E_{x-p_{data}(x)}[logD(x)] + E_{x-p_{z}}[log(1-D(G(z)))]\]
\smallskip

The discriminator is a standard convolutional neural network (CNN) that takes an input image and returns a scalar that represents how real the input image is. There are two convolutional layers that identify 5x5 pixel features and, as with most CNNs, there are fully connected layers at the end. The generator is initialized with a random noise vector while D is trained with a small set of ground truth data. The generator is a deeper neural network, having more convolutional layers and nonlinearities. The noise vector is upsampled and the weights of G are learned through backpropagation, eventually producing data that is classified as real by the discriminator. Further, a key feature of GANs is the ability to produce a larger amount of images than the original dataset.

We utilized this novel architecture to create a pipeline that is able to uphold patient privacy and generate a wider variety of realistic images.

\section{Stage-I GAN}
The purpose of the Stage-I GAN is to generate varied segmentation masks. It is based on the deep convolutional generative adversarial network (DCGAN) architecture [13], and built on the TensorFlow platform. This network has demonstrated competitive results [13] while simultaneously improving training stability in comparison to the standard GAN. The distinctive feature of the DCGAN, compared to other generative models, is that it is fully convolutional, meaning convolutional layers were used instead of pooling layers. Pooling layers reduce the spatial size of the representation, and although they improve computational efficiency, they also result in the loss of important features found in medical images. The generator is initialized with a noise vector, which is fed through multiple strided convolutions to generate a synthetic image.

We used the cross-entropy loss function to train the discriminator in the Stage-I GAN:

\[ l_{D}=\frac{1}{m}\sum_{i=1}^{m}[log(D(G(z^{i}))) + log(1-D(x^{i}))]\]

D is the discriminator, G is the generator, m refers to mini-batch size,  z is the corresponding input noise vector, x is the image, and i is the index of the image. The generator’s loss is described by:

\[l_{G}=\frac{1}{m}\sum_{i=1}^{m}log(1-D(x^{i}))\]

As a result of these two connected loss functions, the generator and discriminator are constantly competing with each other to minimize their respective loss functions. We trained this model on an NVIDIA Tesla K80 GPU.

\section{Stage-II GAN}
The purpose of Stage-II GAN is to translate segmentation masks to corresponding photorealistic images. Stage-II GAN is also built on the TensorFlow platform. Our model is based on an image-to-image translation network proposed by Isola et al. [14] in November 2016; specifically, a vessel-to-retina implementation built by Costa et al. [15].

This network is a special form of GAN known as a conditional generative adversarial network (CGAN). It aims to condition the two networks D and G to a vector y and input image X that represents the mapping between the segmentation mask and photorealistic image. Similar to the regular GAN, the CGAN can be modeled by this function (with the additional input parameter y):

\[min_{G}max_{D}V(D,G) = \]  \[ \mathbb{}{E}_{p_{data}}[logD(x,y)]+  \mathbb{}{E}_{z-p_{z}}[log(1-D(G(z,y))),y]\] \\

The Stage-II GAN is trained with corresponding pairs of real fundi and segmentations masks in order to find a mapping between the two classes of images. Given a segmentation mask, the model will translate the given geometry to a photorealistic medical image.

\section{U-net}
To evaluate the reliability of our synthetic data, we used it to train a u-net segmentation network which creates a segmentation mask given a photorealistic medical image. The u-net architecture, specifically formulated for biomedical images [16], is derived from an autoencoder architecture that relies on unsupervised learning for dimensionality reduction. The u-net is especially useful for biomedical applications since it does not contain fully connected layers, imposing no restriction on input image size and allowing a significantly higher number of feature channels than a regular CNN. Receptive fields after convolution are also concatenated with receptive fields in the decoding process. This allows the network to use original features along with ones after the up-convolution. 

Segmentation is an important task in machine learning used to partition an image into relevant parts. It is also especially useful in medicine to outline malignant bodies and abnormalities such as tumors. When examining retinal images, doctors commonly search for microaneurysms in the blood vessels for the diagnosis of diabetic retinopathy. A u-net segmentation network trained with synthetic data, as shown with our results below, can easily automate and improve accuracy for this process. This is just one of the many applications for  the data produced by our pipeline in computer-aided medical diagnosis.

\section{Evaluation Metrics}
Our pipeline produced synthetic segmentation masks along with corresponding fundi images. We used this data to train a u-net segmentation network. We evaluated the u-net on test images from the DRIVE database and compared them with the ground truth to calculate an F1 score. We also calculated the variance between the synthetic and real datasets through a Kullback–Leibler (KL) divergence score.

When considering GANs, we must analyze the adversarial divergence to calculate the statistical correlation between the generated and original data. The KL divergence score has been the standard to measure this for generative models, calculated by:

\[KL(P,Q)=\sum_{i}^{ } P_{i}(ln\frac{P_{i}}{Q_{i}})\]

We also used the universal F1-score, calculated by taking the harmonic mean of precision and recall. This score can display the similarity between two images, which we use to compare the segmentations produced from our synthetically trained u-net and DRIVE-trained u-net to ground truth segmentations.

\section{Quantitative Results}

We received an F1 accuracy rating of 0.8877 for our synthetically trained u-net and an F1 accuracy of 0.8988 for our DRIVE-trained u-net. The negligible difference between the two scores displays the quality of our produced training data. 

To test for variance, we obtained a KL-divergence score that shows the difference between the distributions of two datasets. The synthetic data score of 4.759 is from comparing the synthetic and real datasets, while the real-data score of 4.212 x 10-4 was measured by comparing two random subsets of the real data. This low score is expected as the two subsets of images are from the same dataset. The synthetic-data score is higher than the real-data score, showing that our synthetic data does not simply copy the original distribution. 


\begin{figure}[ht!]
\centering
\includegraphics[width=70mm]{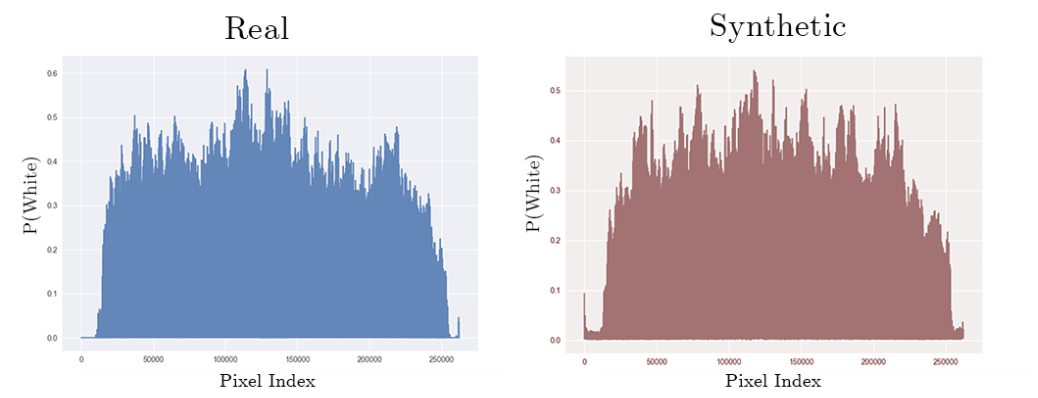}
\caption{Pixel-intensity distribution of real and synthetic datasets.}
\end{figure}

\section{Qualitative Results}

\begin{figure}[ht!]
\centering
\includegraphics[width=70mm]{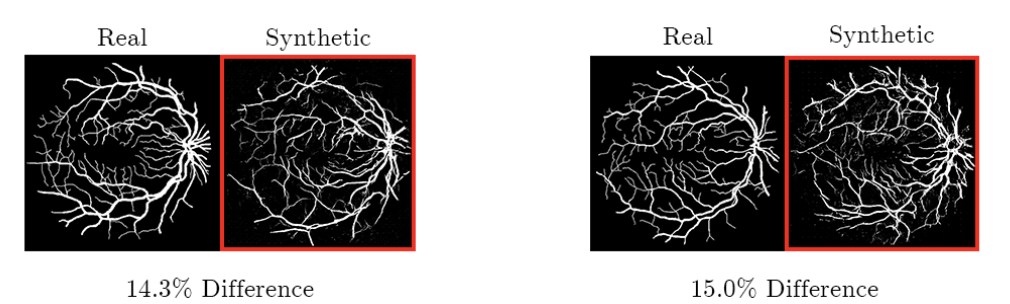}
\caption{Example generated segmentation masks and their closest mask in the training dataset.}
\end{figure}

\begin{figure}[ht!]
\centering
\includegraphics[width=70mm]{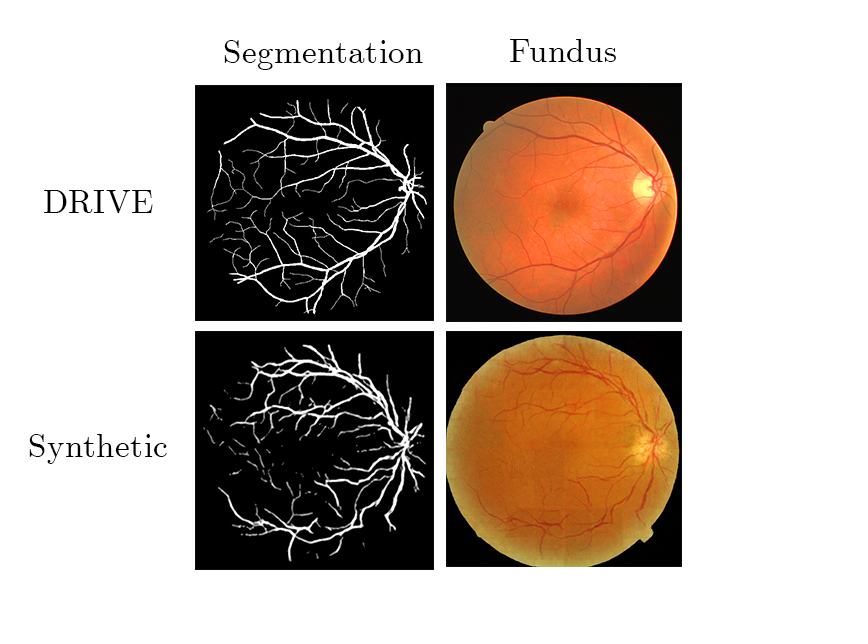}
\caption{Examples from the DRIVE dataset and our synthesized dataset.}
\end{figure}

\section{Pipeline Validation}

To confirm the flexibility of our pipeline, we tested it on a second dataset. Using the  BU-BIL database [17] of 35 rat smooth muscle cell images and segmentations as the training data for our pipeline, we were able to produce a synthetic version of the data.

We chose this database due to its intense variation. The subject in each image is varied in  both shape and position, making it difficult for the GAN to learn which features are relevant. However, through our dual GAN pipeline’s hierarchical generation process, we were able to successfully produce realistic smooth muscle cell images as well as corresponding segmentations. 

As described by our pipeline, we first generated segmentation masks of the smooth muscle cells using Stage-I GAN. We then transferred the segmentations to Stage-II GAN where they were translated into photorealistic smooth muscle cells.

\begin{figure}[!htb]
  \centering
  \includegraphics[width=190pt]{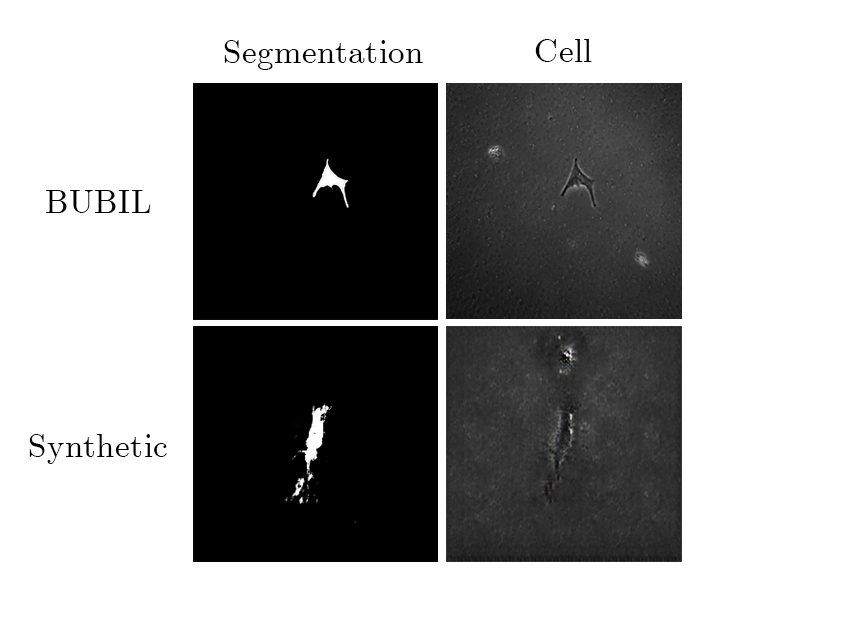}
  \caption{Graphic displaying examples from BUBIL and a
corresponding synthesized dataset.}
  \label{fig:boat1}
\end{figure}

It is important to note this was done on an extremely small dataset of 35 images for both Stage-I and Stage-II to test the limits of our pipeline. The results show that Stage-II was able to learn the correspondence between the segmentation mask and the photorealistic image, but a greater variety of data would be helpful to develop the natural background found in the original images.

\section{Discussion}
Due to the extreme variation of medical imaging data (various illuminations, noise, patterns, etc.), a single GAN is unable to produce a convincing image (see Figure 8). The GAN is unable to determine complex structures, as seen with the poorly defined vessel tree structure and dark spots. It is only able to identify simple features such as general color, shape, and lighting. 

\begin{figure}[!htb]
  \includegraphics[width=100pt]{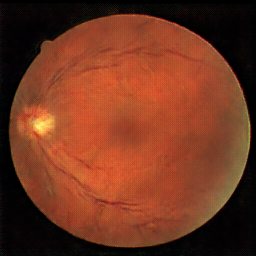}
  \centering
  \caption{Example of a retina image from a single GAN pipeline.}
  \label{fig:boat1}
\end{figure}

This lack of detail is unacceptable for medical image generation, as medical images have many intricacies that must be accurately represented for the data to be usable. Our dual GAN architecture improves the quality of synthetic images by breaking down the challenging task of generating medical images into to hierarchical process. Stacking GANs has been shown to be effective in refining the images produced by a GAN, as seen with Zhang et al [18]. This also allows the unstable nature of GANs to be controlled by providing each GAN with a relatively elementary task. Stage-I GAN focuses only on a much lower dimensional problem: generating unique segmentation geometries, while ignoring photorealism. This allows Stage-II GAN to only generate the colors, lighting, and textures of the medical image from the given geometry. Because the geometry is generated in a lower dimensional image by a separate GAN, an unrealistic vessel geometry causes a larger loss compared to a single GAN that produces unrealistic geometries in its high dimensional fundi images. This system allows both GANs in our pipeline to perform at a high level and reach convergence faster, creating images with more realistic geometries and textures than an ordinary single GAN system.

In addition, the nature of our pipeline produces a wider variety of images than the original dataset. This is because our pipeline generates images that are between the data that formed the distribution. As shown by Figure 5 and Figure 6, our synthetic dataset keeps the general statistical distribution of the real dataset while producing original images. Our pipeline can produce larger quantities of images for effective use in data-driven machine learning tasks, while avoiding legal concerns regarding patient privacy.

\section{Conclusion}
We have proposed a pipeline that is able to generate medical images for a segmentation task end-to-end, using a pair of generative adversarial networks. Our method decomposes the image generation process into two parts: Stage-I GAN which focuses on creating varied geometries of the segmentation mask and Stage-II GAN which transforms the geometry into a photorealistic image. Given a dataset of real images, it can produce larger amounts of synthetic data that is not an image of any real patient, meaning that data produced by our pipeline can be distributed in the public domain. This is a significant step towards the creation of a public and synthetic medical image dataset, analogous to ImageNet. To further this purpose, we have created an online synthetic medical imaging database known as SynthMed. We plan to populate this database with synthetic data from private research.

We hope that future researchers will apply similar synthetic data techniques to provide public access to their private data for the further advancement and development of computer-aided medical diagnosis.

\section{Future Work}
We believe that our pipeline of dual generative adversarial networks can also be applied to fields outside of medical imaging. Specifically, scene generation has been a challenging topic in Computer Vision, due to the complexity and variance of the images. Our two-stage pipeline may be used to simplify the problem where simple features of the scene can be generated using Stage-I GAN and details can be learned through Stage-II. Researchers have shown in the past that single GANs are able to translate manually done photo segmentations to realistic scenes, as seen with Isola et al. in facade generation [14].

We hope to optimize Stage-I in future works by exploring other representations of the segmentation masks. For example, instead of the GAN in Stage-I producing an image, generative models trained with other representations such as bezier curves, 2D point clouds, or skeletons may be used to reduce dimensionality in the generation process. This would reduce computational time as well as the chance of artifacts.

Using our pipeline for different datasets may require the tuning of hyperparameters for increased effectiveness, as is the case with most neural networks. In addition, the development of deeper and more advanced architectures could be implemented to replace certain networks in our pipeline.

Our pipeline relies on a set of accurate data with high variance. For our pipeline to be executed on a variety of medical images, we must have access to private research data. Access to these private collections of images to generate synthetic data is the key to opening up public collaboration for more advanced automated medical image interpretation.

\begin{figure}[!ht]
\includegraphics[width=\linewidth]{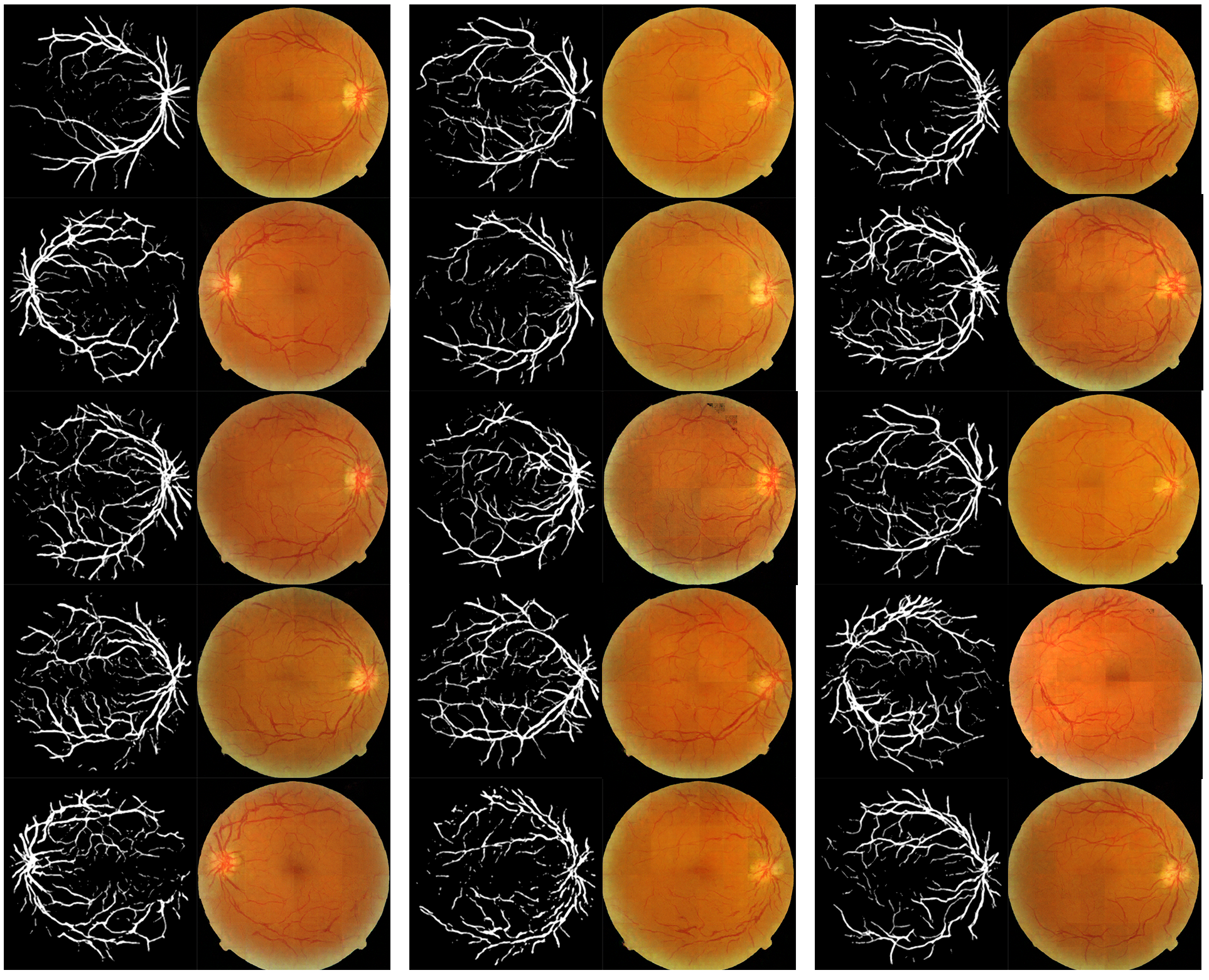}
\caption{Examples of Synthetic Retinal Data}
\end{figure}

\section*{References}
\medskip

\small

[1] Doi, K. (2007). Computer-Aided Diagnosis in Medical Imaging: Historical Review, Current Status and Future Potential. Computerized Medical Imaging and Graphics : The Official Journal of the Computerized Medical Imaging Society, 31(4-5), 198–211. http://doi.org/10.1016/j.compmedimag.2007.02.002

[2] Baris Kayalibay, Grady Jensen, and Patrick van der Smagt. Cnn-based segmentation of medical imaging data. CoRR, abs/1701.03056, 2017. 

[3] Ishida, T, and S Katsuragawa. “[Overview of computer-Aided diagnosis].” Nihon Igaku Hoshasen Gakkai zasshi. Nippon acta radiologica., U.S. National Library of Medicine, July 2002, www.ncbi.nlm.nih.gov/pubmed/12187835

[4] Secretary, HHS Office of the, and Office for Civil Rights (OCR). “Your Rights Under HIPAA.” HHS.gov, US Department of Health and Human Services, 1 Feb. 2017, 
www.hhs.gov/hipaa/for-individuals/guidance-materials-for-consumers/index.html. 

[5] Christopher Cunniff, Janice L.B. Bryne Louanne M. Hudgins, John B. Moeschler, Ann Haskins Olney, Richard M. Pauli, Lauri H. Seaver, Cathy A. Stevens, Christopher Figone. Informed consent for medical photographs, Dysmorphology Subcommittee of the Clinical Practice Committee, American College of Medical Genetics, https://www.acmg.net/staticcontent/staticpages/informed\_consent.pdf

[6] Olga Russakovsky, Jia Deng, Hao Su, Jonathan Krause, Sanjeev Satheesh, Sean Ma, Zhiheng Huang, Andrej Karpathy, Aditya Khosla, Michael S. Bernstein, Alexander C. Berg, and Fei-Fei Li. Imagenet large scale visual recognition challenge. CoRR, abs/1409.0575, 2014. 

[7] Aditi Ramachandran, Lisa Singh, Edward Porter, Frank Nagle. Exploring Re-identification Risks in Public Domains, Georgetown University, Harvard University, https://www.census.gov/srd/CDAR/rrs2012-13\_Exploring\_Re-ident\_Risks.pdf

[8] Jarmin, R. and Louis, T. (2014). [ebook] Washington: U.S. Census Bureau, Center for Economic Studies, https://www2.census.gov/ces/wp/2014/CES-WP-14-10.pdf 

[9] Satkartar K. Kinney, Jerome P. Reiter, Arnold P. Reznek, Javier Miranda, Ron S. Jarmin, and John M. Abowd. Towards Unrestricted Public Use Business Microdata: The Synthetic Longitudinal Business Database. International Statistical Review, 79(3):362–384, December 2011.

[10] J.J. Staal, M.D. Abramoff, M. Niemeijer, M.A. Viergever, and B. van Ginneken. Ridge based vessel segmentation in color images of the retina. IEEE Transactions on Medical Imaging, 23(4):501–509, 2004. 

[11] Decencière et al.. Feedback on a publicly distributed database: the Messidor database. Image Analysis \& Stereology, v. 33, n. 3, p. 231-234, aug. 2014. ISSN 1854-5165. 

[12] I. J. Goodfellow, J. Pouget-Abadie, M. Mirza, B. Xu, D. Warde-Farley, S. Ozair, A. Courville, and Y. Bengio. Generative Adversarial Networks. ArXiv e-prints, June 2014. 

[13] Alec Radford, Luke Metz, and Soumith Chintala. Unsupervised representation learning with deep convolutional generative adversarial networks. CoRR, abs/1511.06434, 2015. 

[14] P. Isola, J.-Y. Zhu, T. Zhou, and A. A. Efros.  Image-to-Image Translation
with Conditional Adversarial Networks. ArXiv e-prints, November 2016

[15] P. Costa, A. Galdran, M. Ines Meyer, M. D. Abramoff, M. Niemeijer, A. M.
Mendonca, and A. Campilho. Towards Adversarial Retinal Image Synthesis. ArXiv e-prints, January 2017

[16] Olaf Ronneberger, Philipp Fischer, and Thomas Brox. U-net: Convolutional networks for biomedical image segmentation. CoRR, abs/1505.04597, 2015. 

[17] D. Gurari, D. Theriault, M. Sameki, B. Isenberg, T. A. Pham, A. Purwada, P. Solski, M. Walker, C. Zhang, J. Y. Wong, and M. Betke. "How to Collect Segmentations for Biomedical Images? A Benchmark Evaluating the Performance of Experts, Crowdsourced Non-Experts, and Algorithms." Winter conference on Applications in Computer Vision (WACV), 8 pp, 2015. [In Press].

[18] P. Costa, A. Galdran, M. Ines Meyer, M. Niemeijer, M. D. Abramoff,  A. M.
Mendonca, and A. Campilho. "End-to-end Adversarial Retinal Image Synthesis."
\end{document}